%% file: main.tex
\pdfoutput=1

\documentclass[runningheads]{llncs}
\usepackage{graphicx}
\usepackage{subfigure}
\usepackage{adjustbox}
\usepackage[font=small,labelfont=bf]{caption}
\usepackage{amsmath,amssymb} 
\usepackage{color}\usepackage[width=122mm,left=12mm,paperwidth=146mm,height=193mm,top=12mm,paperheight=217mm]{geometry}

\input{utilities}

\begin{document}
\pagestyle{headings}
\mainmatter

\title{Deep Image Set Hashing} 

\titlerunning{Deep Image Set Hashing}

\authorrunning{J. Feng, S. Karaman, IH. Jhuo, S.F. Chang}

\author{Jie Feng, Svebor Karaman, I-Hong Jhuo, Shih-Fu Chang}
\institute{Columbia University\\New York, NY, USA\\{\{jf2776, sk4089, sc250\}@columbia.edu}, ihjuo@gmail.com}

\maketitle

\input{00_abstract}
\input{01_introduction}
\input{02_relatedwork}

\input{03_approach}
\input{04_experiments}
\input{05_conclusions}


\section{Acknowledgments}

This research is based upon work supported by the Office
of the Director of National Intelligence (ODNI), Intelligence
Advanced Research Projects Activity (IARPA),
via IARPA R\&D Contract No. 2014-14071600012. The
views and conclusions contained herein are those of the authors
and should not be interpreted as necessarily representing
the official policies or endorsements, either expressed
or implied, of the ODNI, IARPA, or the U.S. Government.
The U.S. Government is authorized to reproduce and distribute
reprints for Governmental purposes notwithstanding
any copyright annotation thereon.

{\small
	\bibliographystyle{ieee}
	\bibliography{main}
}

\end{document}

%% file: utilities.tex
\usepackage{xcolor}

\renewcommand \paragraph[1] {\vspace{0.05cm} \textbf{#1}}

%% file: 00_abstract.tex
\begin{abstract}

In applications involving matching of image sets, the information from multiple images must be effectively exploited to represent each set. State-of-the-art methods use probabilistic distribution or subspace to model a set and use specific distance measure to compare two sets. These methods are slow to compute and not compact to use in a large scale scenario. Learning-based hashing is often used in large scale image retrieval as they provide a compact representation of each sample and the Hamming distance can be used to efficiently compare two samples. However, most hashing methods encode each image separately and discard knowledge that multiple images in the same set represent the same object or person. We investigate the set hashing problem by combining both set representation and hashing in a single deep neural network. An image set is first passed to a CNN module to extract image features, then these features are aggregated using two types of set feature to capture both set specific and database-wide distribution information. The computed set feature is then fed into a multilayer perceptron to learn a compact binary embedding. Triplet loss is used to train the network by forming set similarity relations using class labels. We extensively evaluate our approach on datasets used for image matching and show highly competitive performance compared to state-of-the-art methods.

\end{abstract}

%% file: 01_introduction.tex
\section{Introduction}

With the ubiquity of camera network, ease of imaging and availability of online data, it is fairly easy to capture and access images in the form of a set.
An image set is a collection of unordered images for a target, e.g. an object, a human face, an event etc. Images within a set could exhibit different characteristics about the target, such as different views of an object or a face, images of a scene taken under different lighting conditions, a set of video frames depicting different poses of a human action.
Thus, an image set contains richer information than a single image and is potentially more useful for problems like object or scene classification, face recognition and action analysis.
Many methods \cite{wang2008manifold,wang2012covariance,lu2015multi,yamaguchi1998face,kim2007discriminative} have been proposed to leverage sets as input for matching problem. Most of these methods focus on set modeling and how to compute a proper matching distance between two sets. They usually don't have a compact representation for sets, making it expensive to store the set models in memory and slow to match, thus hard to scale when the number of targets increases.

Learning-based hashing \cite{gong2013iterative,liu2012supervised,14AAAI-Xia,15CVPRW-Lin,wang2016learning} has received a lot of attention in problems like image retrieval. By computing a binary code for each image, much less memory space is needed and Hamming distance could be used to significantly reduce the matching time. Since the hash codes are learned, they are also able to preserve prior known constraints using supervised or semi-supervised learning and achieve good matching accuracy.
Inspired by image based hashing, we consider the problem of hashing for image sets. While in the individual image case, each image is encoded as a binary code, we seek to represent a set as a single binary code regardless of the set size.
This can effectively reduce the complexity of matching two sets without referring to the individual images composing them and can greatly reduce the time and memory cost during matching. Although the benefits of doing set hashing are obvious and appealing, it is not a trivial task: 1) a proper way of representing a set is needed to effectively integrate information from each image; 2) the hashing process should be connected with feature extraction so the image feature is optimized to achieve accurate hash code matching; 3) the method should work with sets of different sizes with the ability to improve performance when bigger sets are used.

\begin{figure}[t]
\begin{center}
   \includegraphics[width=\linewidth]{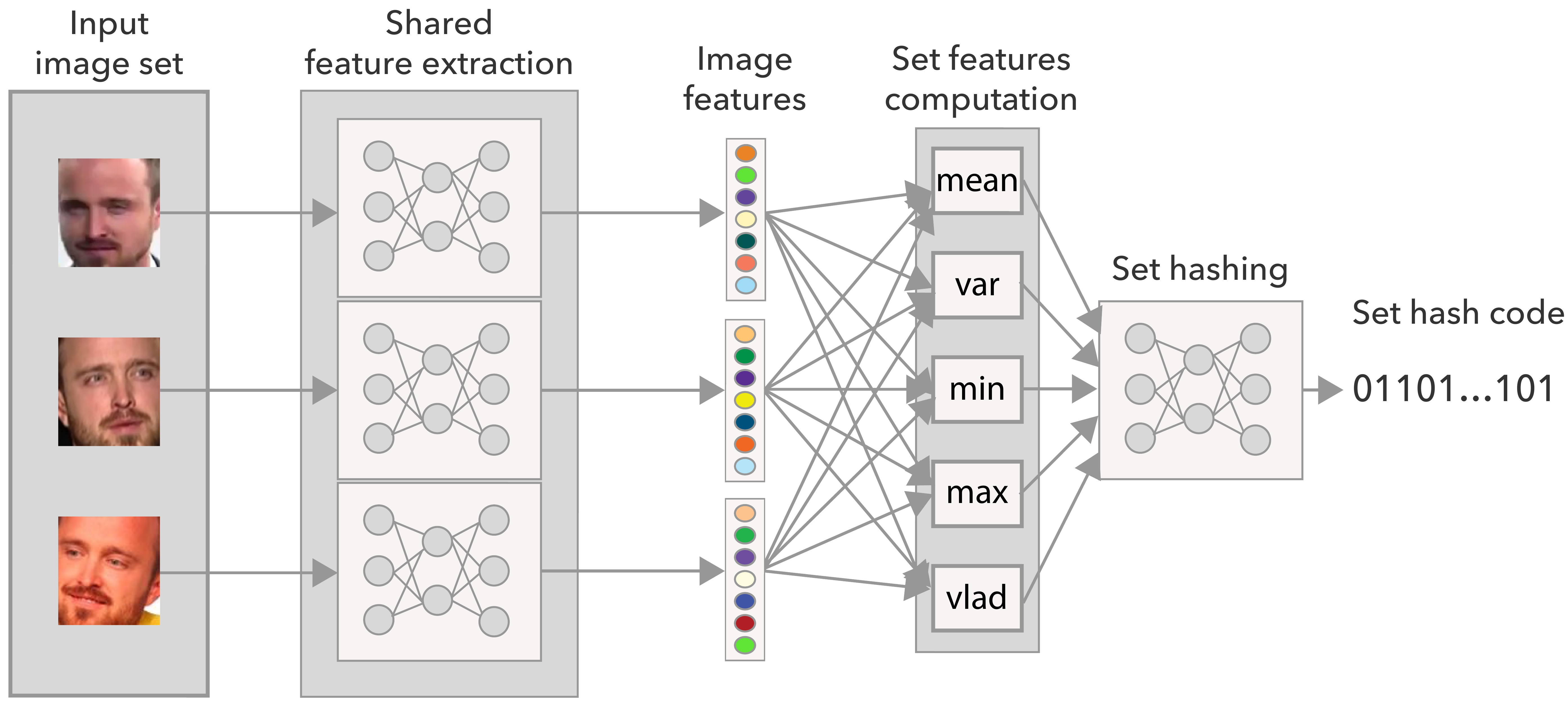}
\end{center}
   \caption{Our deep neural network architecture for set hashing. Each image in the input set goes through a shared feature extraction network producing one feature per image. All features are aggregated in the set feature layer before being encoded to produce a single binary hash code for the whole set.}
\label{fig:overview_network}
\vspace{-2mm}
\end{figure}

In this paper, we propose a novel deep neural network model which takes an image set as input and computes a single compact binary code. Codes are optimized so that sets from the same class have smaller distances while sets from different classes have bigger distances. Each image is first passed through a feature extraction module to get an image feature. Then these features are aggregated to compute a set feature which describes the geometric structure of the features distribution. Finally, the set feature is fed into a multilayer network to be transformed into a binary code. 
An overview of our set hashing network is given in Fig.~\ref{fig:overview_network}. 
To ensure the learned codes are meaningful, the model is trained using triplet loss to get a metric binary embedding by directly optimizing on the distance value. This loss has been shown as an appropriate choice for metric learning~\cite{schroff2015facenet,parkhi2015deep}. 
We evaluate the proposed method on multiple datasets used for image set matching, and show our method compares favorably with regards to the state-of-the-art including both single image hashing methods and set matching.
Our contributions can be summarized as follows:
\begin{enumerate}
	\item We propose a single end-to-end deep neural network structure which learns a binary code for a given image set. The model allows for learning of image features, set feature aggregation, and the final hashing step in a single integrated framework.
	\item We incorporated two types of complementary features to represent a set of image features - one capturing the local distribution geometry and the other capturing encoding of the set using a global dictionary. The combination of the two enables learning high quality set hashing codes.
	\item Our extensive experiments show the proposed set hashing method achieves much better retrieval performance compared to state-of-the-art single image hashing methods and is similarly effective compared to state-of-the-art set matching algorithms but without suffering from the complexity linearly growing with the database size. This makes set based matching scalable
to large datasets.
\end{enumerate}

%% file: 02_relatedwork.tex
\section{Related works}

We will first review here the related literature that studied how to represent and match image sets. Then we discuss works related to deep learning for hashing and set representation.

\subsection{Image set representation and matching}

The key problems for set matching are how to represent a set of images and how to properly compute the similarity between two given sets. The two being linked as the appropriate similarity measure depends on the set representation at hand. The existing works can be grouped into two categories: parametric and non-parametric methods.

Parametric approaches seek to represent the distribution within a set, for example using a single Gaussian~\cite{shakhnarovich2002face} or Gaussian Mixture Models (GMM)~\cite{arandjelovic2005face}
or as probability distribution functions (PDFs) using kernel density estimators~\cite{harandi2015beyond}. The distribution representations are compared with f-divergence functions such as the Kullback-Liebler (KL) divergence. These approaches need to estimate a complex model with potentially many parameters for each set. Moreover they make strong assumption on the distribution within one set, and between gallery and query sets causing them to be sensitive to statistical noise.

The non-parametric methods seek a  more effective set representation.
Several works have focused on finding some representative exemplars: mean~\cite{wang2008manifold},  affine hull or convex hull~\cite{cevikalp2010face}, approximated nearest neighbors~\cite{hu2011sparse}, and regularized nearest points~\cite{yang2013face}. Other works use a geometric representation such as linear subspaces~\cite{hamm2008grassmann,kim2007discriminative} compared with the principal angles method~\cite{kim2007boosted} or the projection kernel metric~\cite{hamm2008grassmann}.
Subspace based methods perform well when each set represents a dense sampling, but tends to struggle when the set is of small size with complex data variations. 
 
Linear subspace models are usually estimated from an eigen-decomposition of the covariance matrix but discarding non leading eigenvectors. Hence, the covariance matrix characterizes the set structure more faithfully and has therefore also been used as a set representation~\cite{wang2012covariance} but its higher dimensionality is a burden when dealing with large scale problems. In the Hashing across Euclidean space and Riemannian manifold (HER) framework proposed in~\cite{li2015face}, video clips are encoded as covariance matrices and embedded into a Reproducing Kernel Hilbert Space (RKHS) before a hash learning procedure relying on SVMs.

\subsection{Deep learning for single image hashing}

In the last couple of years, several approaches were proposed for binary embedding with deep networks. A two stage approach was developed in~\cite{14AAAI-Xia} where the authors first learn target codes using pair similarity supervision and then train a deep hash network to map each sample to its target code. The pre-training procedure renders the method not scalable.
End-to-end deep hash methods were proposed in~\cite{15CVPRW-Lin,lai2015simultaneous,15CVPR-DSR-Zhao}. 
The authors of~\cite{15CVPRW-Lin} proposed to add a latent hash layer to a standard AlexNet and use a classification loss to train their model. If our goal is to learn hash codes for fast retrieval of similar samples, the optimization should focus on learning hash codes for which the hamming distance preserves similarity ranking. The classification optimization of the previous work has no guarantee to do so, hence recent methods prefer to rely on a triplet loss to enforce hamming distances between hash codes to follow relations of the class semantics~\cite{15CVPR-DSR-Zhao}. The authors of~\cite{lai2015simultaneous} also proposed a divide and encode approach to improve bit independence. All these methods proposed solely deep hashing methods that produce a hash code for each single image separately. 

\subsection{Deep learning for set classification}
\label{sec:rw-deepset}
A few recent works have addressed the problem of set classification with deep networks, mainly using multiple class specific models. The authors of~\cite{lu2015multi} propose a multi-manifold deep metric learning approach for image set classification. Class-specific neural networks are trained using a maximal manifold margin that minimizes the intra-manifold and maximizes the inter-manifold distances. In~\cite{hayat2014learning}, class specific auto-encoders are trained and the classification procedure seeks the minimal reconstruction error when auto-encoding each sample of the test set with each class model, the set label being estimated as the most recurring label amongst all images of the test set. A similar classification strategy is used in~\cite{shah2016iterative} where class-specific neural networks are trained using a Pooled Convolutional representation as input.

In these methods, the classification of one test set requires passing each test image through each class network. Hence, the complexity of these methods grows linearly with the number of classes making them intractable for problems with large number of classes in the gallery. Also the test set structure is only used in aggregation of sample distances or classification results. 

\bigskip

To the best of our knowledge, our method is the first to tackle the set hashing problem in an end-to-end learning framework. We specifically aim for a single model that can efficiently encode sets of images from a number of classes without relying on class-specific models that will prevent efficient scalability when the number of target classes grows. Our model enables large-scale set based retrieval thanks to the binary representation.



%% file: 03_approach.tex
\section{Deep Set Hashing Network}

We design a deep neural network model to process input in the form of image sets.
The network structure is composed of three blocks: an image feature extraction phase 
that is applied to each sample, an aggregation layer applied to all the image features within an input set to  compute set features, and finally a hashing phase that encodes the set features into
 a single hash code. The network structure is illustrated in Figure~\ref{fig:overview_network}. 
We detail in this section each building block of our network.

\subsection{Image feature extraction} 
Each image $I_i$ in the input set $\mathcal{S}$ first goes through a feature extraction module to transform the original pixels to a powerful feature representation. 
Convolutional Neural Networks (CNN) are used as the feature extractor given its proven success on various computer vision problems. 
As has been showed in various published works, the CNN structure needs to be engineered to work on specific tasks and data domains. 

To make the image feature relevant and useful, we could choose a network design that has been shown to work well for a particular problem. VGG-16~\cite{simonyan2014very}, Inception network~\cite{szegedy2015going} and the recent ImageNet-winner Residual network~\cite{he2015deep} achieve excellent performance on object classification problem. They would be the preferred architectures for feature extraction when we work with object related datasets.
For face recognition, CASIA network~\cite{yi2014learning}, Google FaceNet~\cite{schroff2015facenet} and Oxford VGG-face~\cite{parkhi2015deep} are some successful examples. Once the feature extraction part is chosen, it is shared among all images. In Section~\ref{sec:train-net}, we will also discuss how the network parameters (e.g., coefficients) will be learned in an end-to-end manner to derive the optimal set hashing method.

\subsection{Set feature computation}

\begin{figure}
	\begin{center}
		\includegraphics[width=\linewidth]{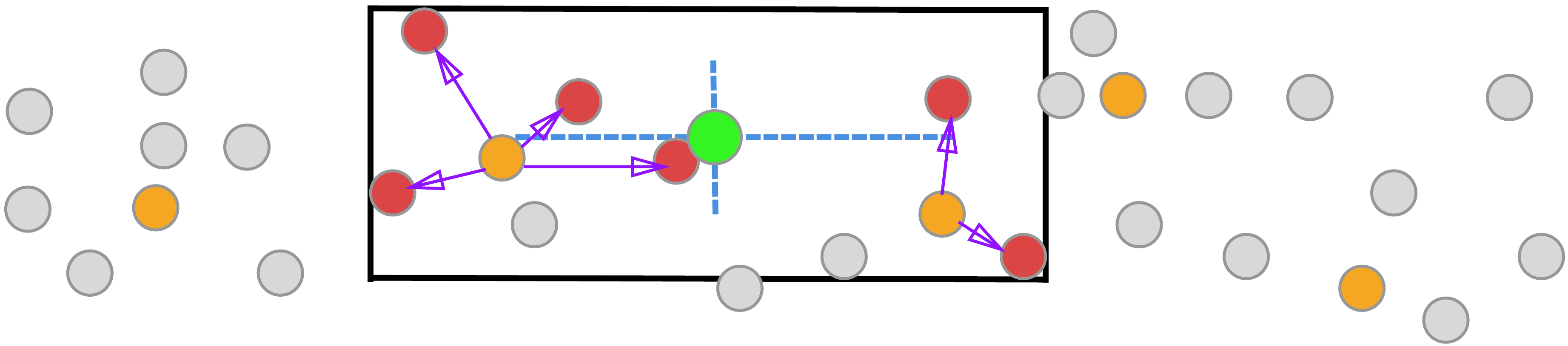}
	\end{center}
	\caption{Illustration of set features in the image feature space: red nodes are images of the set $\mathcal{S}$ considered, gray node are images not in the set, orange nodes are cluster centers of the VLAD dictionary, the green node is $mean(\mathcal{S})$, black lines represent the space delimited by $min(\mathcal{S})$ and $max(\mathcal{S})$, blue lines represent $var(\mathcal{S})$ and purple arrows represent $VLAD(\mathcal{S})$ (best viewed in color). Please see text for more details.}
\label{fig:set_features}
\end{figure}

Given a set of input samples, we need to find a way to merge them into a unified set representation regardless of how many samples are present for a given set. This process enables us to model a set compactly and perform distance measure without resorting to the original images. Meanwhile, information from different images should be captured in the final representation so we benefit from having more data than a single image.

To ensure the set feature is generic to maintain characteristics of the unknown underlying image feature distribution, we derive statistics to describe the global geometry property of the set distribution.  Specifically, for each input set $\mathcal{S}$ we compute:
\begin{itemize}
\item the average feature: $mean(\mathcal{S}) \in \mathbb{R}^d$;
\item the variance of each feature dimension: $var(\mathcal{S}) \in \mathbb{R}^d$;
\item the element-wise minimum feature: $min(\mathcal{S}) \in \mathbb{R}^d$;
\item the element-wise maximum feature: $max(\mathcal{S}) \in \mathbb{R}^d$.
\end{itemize}
$mean(\mathcal{S})$ indicates the average location of the set in the sample feature space. 
$var(\mathcal{S})$ shows how values of each feature dimension changes relative to its mean within the given set. $min(\mathcal{S})$ and $max(\mathcal{S})$ give the range of the set features. The last three measures capture the shape of the set in the feature space.
An illustration of the set statistic feature is shown in Fig~\ref{fig:set_features}. Here each gray node represents an image in the feature space, red nodes denote a set of images. Green node is the mean of the set, the lines surrounding the set gives $min(S)$ and $max(S)$. $var(S)$ describes the average variation of the set around the mean.  From the graph, we can see these summary statistics are useful to provide an overall view of the set, but they only depict the set itself independent of other sets. 

For an effective comparison, it is necessary to derive features based on the overall feature distribution across all set samples. Inspired by the popular pooling methods of image local descriptors in image retrieval such as Bag-of-words (BoW)~\cite{sivic2003video} and its improved derivations, we add another set feature using the Vector of Locally Aggregated Descriptors (VLAD)~\cite{jegou2010aggregating} representation which has been shown very effective in describing a set of features with a lower dimension compared to Fisher Vectors (FV)~\cite{perronnin2007fisher}. In our case, we use the training image pool to construct the dictionary $D$ that is required in computing VLAD features. This dictionary no longer encodes local image patches but the higher level concepts at the image level.
Given the deep features extracted from the CNNs, the dictionary can be used to represent semantic structure like pose, attributes, category etc. To compute the dictionary, we first extract CNN features for all the individual images in the training image sets, we then perform K-Means to get the cluster centroids: 
$\mathcal{C} = \{c_1, \ldots, c_k\}$. 
Assuming a set $S$ of $N$ images $\{I_1, \ldots, I_N\}$ with corresponding CNN features $\{x_1, \ldots, x_N\}$, each component $v_{k,j}$ of the VLAD descriptor for $\mathcal{S}$ is calculated as:

\begin{equation}
v_{k,j} = \sum_{x_i \text{ such that } NN(x_i)=c_k} x_{i,j} - c_{k,j},
\end{equation}
where $k \in [1,K]$ is the cluster index in $D$ and $j \in [1,d]$ indexes the dimension of the CNN image feature $x$.  
L2-normalization is applied to get the final VLAD feature $VLAD(S)$.
As shown in Fig.~\ref{fig:set_features}, orange nodes denote cluster centers and purple arrows denote the difference vector between each set node and its closest cluster center, we can see VLAD describes the distribution pattern of set samples relative to the whole image feature space, thus giving additional information on the set.

The output of the aggregation layer for a set $\mathcal{S}$ is then the concatenation of the five set features $F(\mathcal{S})=(mean(\mathcal{S}), var(\mathcal{S}), min(\mathcal{S}), max(\mathcal{S}), VLAD(\mathcal{S}))$. Notice unlike other existing methods where set feature and image feature are usually computed independently, in our case, by connecting image feature extraction with set feature computation, the image feature will be learned through the end-to-end back propagation training process described in Section~\ref{sec:train-net} below, to optimize the set feature, thus no fixed assumption of the set distribution is made.

\subsection{Set Hashing}
Hashing is performed by multiple fully connected layers with non-linear activation.
For all layers except the last layer, rectified linear unit (Relu) is used while the sigmoid activation is used in the last layer to get output value between $0$ and $1$.
We use $W^i$ to indicate the parameter of $i$th layer and $b^i$ is the corresponding bias.
Given a set aggregate feature $F(\mathcal{S})$, the output of the first layer could be written as: $h_1(F(\mathcal{S}))=s(W^1 F(\mathcal{S})+b^1)$ where $s(.)$ is a nonlinear activation function. Similarly, output of the $i$th layer is: $h_i(F(\mathcal{S}))=s(W^i h_{i-1}(F(\mathcal{S}))+b^i)$. Note that each layer $h_i$ is a multi-dimensional vector. Assuming we use $M$ layers for hashing, our set hashing network output is:
\begin{equation}
H(F(\mathcal{S})) = h_M(F(\mathcal{S})) = s(W^M h_{M-1}(F(\mathcal{S}))+b^M)
\end{equation}
To generate the final binary code, we use $0.5$ to threshold this output:
$B(\mathcal{S}) = sgn(H(F(\mathcal{S}))>0.5)$.

\subsection{Training the network}
\label{sec:train-net}
One of the distinct novelty of our proposed approach is the end-to-end architecture that allows us to learn different parts (image feature, set aggregation, and hashing) in the same model.
To train our network, we need to define a proper loss function given set training data and labels.
In our case, we want the final binary code to preserve set relations defined by corresponding labels, i.e. codes of sets with the same labels should have a smaller Hamming distance while codes of sets with different labels should have a bigger Hamming distance. To train such objective, the triplet loss could be used as in \cite{schroff2015facenet} and \cite{parkhi2015deep}. 

Let's denote a set triplet $t_i$ as $<\mathcal{S}_a, \mathcal{S}_p, \mathcal{S}_n>$ with corresponding feature for each set as $F(\mathcal{S}_a)$, $F(\mathcal{S}_p)$ and $F(\mathcal{S}_n)$. Here $\mathcal{S}_a$ is the anchor set, $\mathcal{S}_p$ is the positive set which has the same class label as $\mathcal{S}_a$, $\mathcal{S}_n$ is the negative set with different class label as $\mathcal{S}_a$. To make the notation simpler, we replace $F(\mathcal{S}_a)$ by $F_a$, $F(\mathcal{S}_p)$ by $F_p$ and $F(\mathcal{S}_n)$ by $F_n$. We get the output from the final hash layer and compute the loss for each triplet:
\begin{equation}
L_{t_i} = max \{0, || H(F_a)-H(F_p) ||_2^2 - || H(F_a)-H(F_n) ||_2^2 + \alpha \}
\end{equation}
where $|| . ||_2^2$ is the squared Euclidean distance.
The overall triplet loss then could be computed as the sum of each individual loss:
\begin{equation}
J_0 = \frac{1}{|\mathcal{T}|} \sum_{t_i \in \mathcal{T}} L_{t_i}
\end{equation}
Here $\mathcal{T}$ is the collection of all training triplets.
Since the hard binary output are non-differentiable, we use the real valued outputs from the sigmoid activation.
To help learn better codes, we introduce two additional cost. First, we compute a distance between real valued output from last hash layer and binary output:
\begin{equation}
J_1 = \frac{1}{2} || \mathbf{B}-\mathbf{H} ||_F^2	
\end{equation}
where $\mathbf{B}$ is the matrix of binary outputs for all input samples, $\mathbf{H}$ is the real valued outputs matrix and $|| . ||_F^2$ is the squared Frobenius norm. This cost helps push the real valued output to be close to binary output.
Second, we would like the learned codes to be balanced, meaning the variance of the output values should be maximized:
\begin{equation}
J_2 = \frac{1}{N} tr(\mathbf{HH}^T)	
\end{equation}
Here $N$ is the total number of samples, i.e. $N = 3 |\mathcal{T}|$.
Overall, our loss function is defined as:
\begin{equation*}
\label{eq:loss_function}
\begin{aligned}
J &= J_0 + \lambda_1 J_1 - \lambda_2 J_2 \\
\end{aligned}
\end{equation*}
$\lambda_1$ and $\lambda_2$ are parameters to balance the different cost terms, they were set empirically to $1$ and $0.1$ for all experiments based on preliminary evaluation.
The optimization is carried out with Stochastic Gradient Descent (SGD). Since our architecture is actually a single network, the back-propagation will optimize the parameters in both the set hashing layers and the image feature extraction layers. The set features are parameter-free and allow the gradient to be back-propagated to the image feature extraction block. 


\subsection{Dealing with varying set size}
\label{sec:set_gen}

\begin{figure}
\centering
\newcommand{\augfigsize}{.175}
\subfigure[Input image]{
\includegraphics[width=\augfigsize\textwidth]{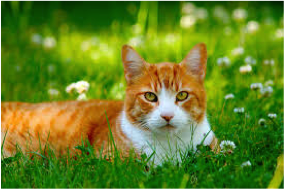}
}
\subfigure[Cropping]{
\includegraphics[width=\augfigsize\textwidth]{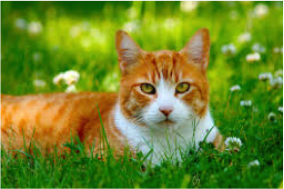}
}
\subfigure[Scaling]{
\includegraphics[width=\augfigsize\textwidth]{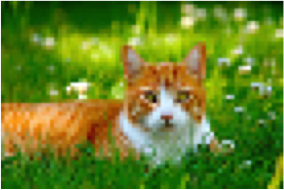}
}
\subfigure[Flip]{
\includegraphics[width=\augfigsize\textwidth]{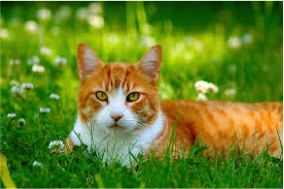}
}
\subfigure[Color shuffle]{
\includegraphics[width=\augfigsize\textwidth]{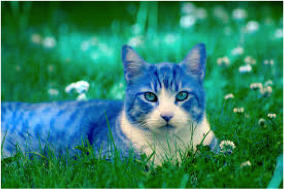}
}
\caption{Examples of augmented images.}
\label{fig:aug_samples}
\end{figure}

Set inputs to our network could have very different sizes. 
When the set size is small, our set feature, especially statistics like mean and variance will not be meaningful. In this case, it is necessary to augment the set to include more images. 
The specific threshold for the set size in deciding whether data augmentation is needed is not important since conceptually more data will help.
We employ 4 types of augmentation to introduce variations: 1) cropping: use a fixed size window to generate random cropped images; 2) scaling: downsample the image to a lower resolution then resize it to the original size; 3) flip: transform the image with a vertical or horizontal symmetry; 4) color shuffle: shuffle color channels to generate different looking versions. Example images are shown in Fig~\ref{fig:aug_samples}. These types can be combined to generate many extra images so we can have a reasonably big set. This augmentation process brings more variation to the training process, making the learned model more robust to changes and noise present in the data. Besides the issue of small sets, it is possible that we don't have a sufficient number of sets for training, in this case, random subsets are sampled from the original set to expand the training data. The generated subset shares the same set label. 

%% file: 04_experiments.tex
\section{Experiments}

To evaluate our proposed set hashing method, we perform various types of experiments. First, 
we compare retrieval performance between our set hashing and different traditional individual image hashing techniques to show that our method can effectively use set input to achieve much better performance.
Second, we compare with state-of-the-art set matching algorithms and show our learned set codes can achieve similar performance with a much smaller memory footprint for the representation and matching complexity. Third, we evaluate the influence of some important parameters of our network.

\subsection{Comparison with individual image hashing}
\label{sec:comp_img_hashing}

\begin{table}
\centering
\setlength\tabcolsep{4pt}
 \begin{adjustbox}{max width=0.9\textwidth}
\begin{tabular}{|c||c|c|}
		\hline
		Methods & \textbf{MNIST} (32bits) & \textbf{CIFAR10} (32bits) \\ 
		\hline
		ITQ~\cite{gong2013iterative} & $0.82$ & $0.21$ \\ \hline
		ITQ (deep feature) & 0.97 & 0.49 \\ \hline
		KSH~\cite{liu2012supervised} & $0.98$ & $0.30$ \\ \hline
		KSH (deep feature) & 0.97 & 0.59 \\ \hline
		Deep image hashing~\cite{xia2014supervised} & 0.97 & 0.52 \\ \hline
		Deep image hashing~\cite{lin2015deep} & $0.98$ & $0.64$ \\ \hline
		Our set hashing & \textbf{0.99} & \textbf{0.81} \\ \hline
\end{tabular}
\end{adjustbox}
\vspace{2mm}
\caption{Mean Average Precision (MAP) of different hashing methods on MNIST and CIFAR10.}\label{table:mapmnistcifar10}
\end{table}

%
Two standard hashing datasets are used: MNIST and CIFAR10. MNIST~\cite{lecun1998mnist} consists of 70000 handwritten digit images from 10 classes (0-9). Each image is grayscale of size $28$x$28$ pixels. 50000 images are used as training data and the rest 20000 images are used as testing data. CIFAR10~\cite{krizhevsky2009learning} is a subset of the Tiny image dataset~\cite{torralba200880}. It contains 10 object classes, with a total 60000 images. Each image is $32$x$32$ pixels in color. We use 50000 images for training and 10000 for testing. Since these two datasets have no set information, we construct our own sets. For each image in the training data, we randomly sample additional images to form a set of size 10. This ensures each image appears at least once in training sets. This is also used to form query sets of size 30.

For image based hashing, the performance is measured by computing set distance as the mean average image distance. 
The distance between the query set and each target set is computed in the following way: first, each image in the query set is used as a query and the average Hamming distance of its hash code and the hash codes of images in a target set is computed, and then the mean of such average distance is computed over all images in the query set. 
This involves many Hamming distance computations. For set hashing, Hamming distance between two set codes are computed directly between there set binary codes hence requiring a single Hamming distance computation. During the training process, only the training set is used. During evaluation, testing data is used as query and training data is used as gallery (namely target database) to perform retrieval. Mean Average Precision (MAP) is used as the performance evaluation metric. Each image set in the test set is used as a query and when the retrieved image set belong to the same class, the result is considered correct. 

If not specified otherwise, raw pixels are taken as input to learn the codes. The deep features used by ITQ and KSH are extracted from CNNs trained on each dataset for classification using the same structure as the feature extraction part of our model. Details about the CNN architectures used for each dataset are given in Section~\ref{sec:implem}.  We use two layers for our set hashing, with 512 and 32 nodes respectively, which hence produce 32 bits codes. 

Comparisons are done with both standard hashing methods and deep learning based hashing methods, results are shown in Table~\ref{table:mapmnistcifar10}. Given that MNIST is a simple dataset which contains limited variation for each character, all methods perform quite well with deep learning based approaches reaching very high MAP value close to $1$. Our set hashing is even able to improve slightly to $0.99$. 
On CIFAR10, all methods perform lower than on MNIST since CIFAR10 is more challenging with low quality images of objects under large appearance variation.  KSH and ITQ trained directly on pixels perform poorly. By using deep features, their performance is greatly improved. However, an end-to-end feature learning and hashing network for image~\cite{lin2015deep} achieves a higher MAP of $0.64$.
Our set based hashing method further increases the performance beating image based hashing methods by a large margin, indicating the effectiveness of our method in encoding multiple images within a set into a single binary code.
 
\subsection{Comparison with set matching}

We use some of the most popular image set data to compare with non-hashing based set matching methods.
Two face datasets are used, including a small scale Honda/UCSD dataset~\cite{lee2003video} and the larger scale YouTube Celebrity~\cite{kim2008face}. For object set matching, the ETH-80 dataset~\cite{leibe2003analyzing} is used.

The Honda/UCSD dataset contains a total of $59$ video sequences from $20$ different subjects. Each video has a frame count ranging from $12$ to $645$. The original face images have a resolution of $32$x$32$ pixels which are very low quality, they are resized to $100$x$100$ to be fed into our face feature extraction neural network, see Table~\ref{tab:casia_network}.

YouTube Celebrities (YTC) dataset consists of $1910$ videos of $47$ celebrities. The face images exhibit large variations in illumination, pose and expression. The face region of each frame is extracted using an off-the-shelf face detector~\cite{viola2001rapid}. For each person, three videos are randomly selected as the training data (which is used as the gallery during testing), the remaining videos are used as test queries. 

The ETH-80 object dataset contains image sets of $8$ object categories: apples, cars, cows, cups, dogs, horses, pears and tomatoes. Each category has ten object instances and each instance has images under 41 different orientations. We use $5$ instances as training data (which become gallery sets during testing), and the remaining $5$ are used as test query sets.

\begin{table}
	\centering
	\setlength\tabcolsep{4pt}
	\begin{adjustbox}{max width=0.9\textwidth}
		\begin{tabular}{| l || c | c | c | c |}
			\hline
			Methods &  \textbf{Honda/UCSD} &  \textbf{YTC} &  \textbf{ETH-80} \\ 
			\hline
			DCC~\cite{kim2007discriminative} & 92.56 & 51.42 & 91.75 \\ \hline
			MMD~\cite{wang2008manifold} & 92.05 & 54.04 & 77.50 \\  \hline
			AHISD~\cite{cevikalp2010face} & 91.28 & 61.49 & 78.75 \\ \hline
			CHISD~\cite{cevikalp2010face} & 93.62 & 60.42 & 79.53 \\ \hline
			SANP~\cite{hu2011sparse} & 95.13 & 65.60 & 77.75 \\ \hline
			CDL~\cite{wang2012covariance} & 98.97 & 56.38 & 77.75 \\ \hline
			RNP~\cite{yang2013face} & 95.90 & 65.82 & 96.23 \\ \hline
			ADNT~\cite{hayat2014learning} & 100 & 71.35 & 98.12 \\ \hline
			RT~\cite{hayat2014reverse} & 100 & 74.10 & 95.50 \\ \hline
			IDLM~\cite{shah2016iterative} & 100 & \textbf{76.52} & \textbf{98.64} \\ \hline
			Ours (512 bits) & \textbf{100} & 75.03 & 97.23 \\ \hline
		\end{tabular}
	\end{adjustbox}
	\vspace{2mm}
	\caption{Average classification accuracy on image set datasets, compared with state-of-the-art image matching techniques.}\label{table:image_set_compare}
\end{table}

For evaluation, we use classification accuracy as metric to be comparable with both set matching and classification approaches. The nearest neighbor is used to predict the label. 
In this experiment, we use two hashing layers with 1024 and 512 nodes.
This configuration works best across all datasets.
We summarize the results in Table~\ref{table:image_set_compare}.
All recently proposed methods~\cite{hayat2014learning,hayat2014reverse,shah2016iterative} perform well on these datasets. On Honda/UCSD dataset, four methods are able to achieve perfect classification including ours. Our method performs slightly worse on ETH-80 than ADNT and IDLM but still much better than all other methods with only 512 bits for each set. 
YTC is more challenging given face images captured in vastly difference conditions. Our method is able to perform second best with a little lower accuracy than IDLM. We would like to point out that, as discussed in Section~\ref{sec:rw-deepset}, both IDLM and ADNT train deep models for each class specifically which is more accurate without interference of samples from other classes. Meanwhile, it is more expensive to compute with the complexity in training and testing linearly increasing according to the number of classes. Each image in the set has to be processed independently and voting is needed to decide the final set label. Our method uses the same binary coding representation for all classes and regardless of how many images are presented in each set which hence does not need to grow as the number of class increases and is much faster to compute. 
%

\subsection{Effects of key configurations}
There are several important configurations in our network which will influence the final output. Here we conduct a detailed analysis on each factor. 
To do that, we use CIFAR10 as a testbed given it is easy to form sets and has enough data to train our model from scratch.
Similar settings are used as in Section~\ref{sec:comp_img_hashing} with modifications applied to each specific task. The results are reported by averaging 10 trials of random set generation.
For set hashing, we are interested in the following questions: 1) how different set sizes will affect our method; 2) how effective is the augmentation when applied to small sets; 3) how useful each type of set feature is.

\subsubsection{Influence of set size}
To evaluate the effect of different set sizes, we construct query sets with sizes ranging from $5$ to $75$ while keeping the gallery set sizes unchanged. 
The MAPs for each set size are shown as the red curve in Fig.~\ref{fig:res_set_aug}. Each circle denotes the set size used. It is clear that our performance increases consistently when the set size increases which indicates our method is able to leverage more data to improve its performance. As the set size goes beyond $50$, the performance changes slowly and tend to converge around $0.85$ which hints $50$ might be a sufficient number of samples in a set for this dataset.

The other factor we want to verify is how data augmentation helps in our case. Data augmentation is used to generate larger sets from smaller sets by applying transformations on the initial images of a set as described in Section~\ref{sec:set_gen}. We conduct an experiment by augmenting a query set and compare the matching performance with a query set of the same size without augmentation. The result is shown as the blue curve in Fig.~\ref{fig:res_set_aug}. Each blue circle is generated by taking the previous set size and augmenting the sets to the current target set size, e.g. the blue circle at set size $10$ is the performance evaluated when using initial query sets of size 5 and produce 5  additional images through augmentation to form a set of size 10. From the graph we can see the MAP of set sizes by augmentation is very similar with sets without augmentation. This shows our augmentation works quite well to add additional query information helping the matching process.

\vspace{4mm}
\begin{minipage}{\textwidth}
  \begin{minipage}[b]{0.5\textwidth}
  \includegraphics[width=0.85\textwidth]{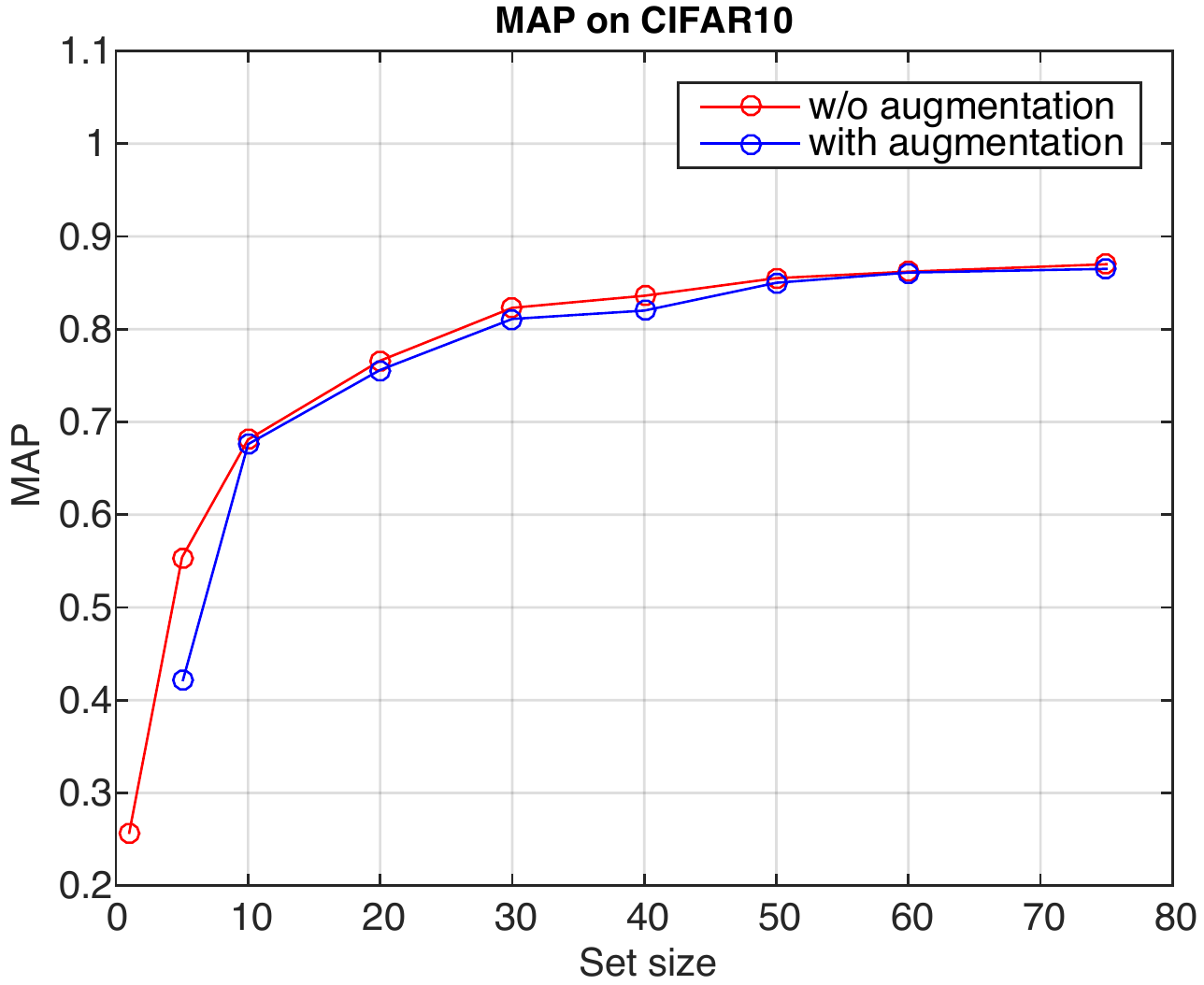} 
  \captionof{figure}{Evaluation of the impact of set sizes and data augmentation.
  \label{fig:res_set_aug}} 
  \end{minipage}
  \hspace{2mm}
   \begin{minipage}[b]{0.4\textwidth}
       \centering
	\setlength\tabcolsep{4pt}
		\begin{tabular}{c|c}
			\hline
			\textbf{Set feature} & MAP \\ 
			\hline
			all statistics & 0.776 \\ 
			VLAD & 0.765 \\
			all features & \textbf{0.797} \\
			\hline
		\end{tabular}
		\captionof{table}{Mean Average Precision (MAP) of different set features on CIFAR10 dataset.\label{table:map_cifar10_setfeature}}
 \end{minipage}
  \hfill
\end{minipage}



\subsubsection{Set feature comparison}
We evaluate the performance using each type of set features (VLAD and statistics like mean and variance etc) and their combination.
The detailed results are shown in Table~\ref{table:map_cifar10_setfeature}. By only using one of the set features, our method is able to get reasonably good results with statistics performing better than VLAD feature in this particular case. However, after combining these two features, the result is improved showing the complementary property of these two set features, they capture somehow different characteristics of a set in a local (set itself) and global (among all sets) context. 

\subsection{Implementation details}
\label{sec:implem}

Based on the target datasets, we use different pre-trained CNN model as feature extraction module. For experiments on face dataset, we use the CASIA face model~\cite{yi2014learning} whose architecture is shown in Table~\ref{tab:casia_network}. The output of pool5 layer is treated as feature. We replaced Relu with parametric Relu as activation function which gives a little improvement in performance. 
For the object dataset, we use the VGG-16~\cite{simonyan2014very} model which won ILSVRC-2014 competition~\cite{russakovsky2015imagenet}. For MNIST and CIFAR10, we use network structures defined in Table~\ref{tab:mnist-net} and Table~\ref{tab:cifar-net} respectively.
For all the datasets except MNIST and CIFAR10, we expand each original image set by generating 50 subsets. If the set is small, we first augment using random transformations as described in Section~\ref{sec:set_gen} to make the set at least 10 images, then do the sampling. Training process is done by generating 3000 triplets in each epoch from training image sets and 10000 epochs are used. The triplet margin $\alpha$ is set to $\frac{1}{2}\sqrt{b}$ where $b$ is the number of bits used in a given experiment, this setting makes the margin grow sub-linearly with an increasing binary codes size. The deep network is implemented in Theano~\cite{Bastien-Theano-2012} and Lasagne~\cite{sander_dieleman_2015_27878}. Training is performed on Nvidia GTX 980 GPU.

\begin{table}
 \centering
 \setlength\tabcolsep{3pt}
 \begin{minipage}{0.47\textwidth}
 \centering
 \begin{adjustbox}{max width=0.99\textwidth}
 \begin{tabular}{| l | l | c | c |}
 \hline
 Name & Type & Filter size / Stride & Output size \\ \hline
 Conv11 & Convolution & 3x3 / 1 & 100x100x32 \\ \hline
 Conv12 & Convolution & 3x3 / 1 & 100x100x32 \\ \hline
 Pool1 & MaxPooling & 2x2 / 2 & 50x50x64 \\ \hline
 Conv21 & Convolution & 3x3 / 1 & 50x50x64 \\ \hline
 Conv22 & Convolution & 3x3 / 1 & 50x50x128 \\ \hline
 Pool2 & MaxPooling & 2x2 / 2 & 25x25x128 \\ \hline
 Conv31 & Convolution & 3x3 / 1 & 25x25x96 \\ \hline
 Conv32 & Convolution & 3x3 / 1 & 25x25x192 \\ \hline
 Pool3 & MaxPooling & 2x2 / 2 & 13x13x192 \\ \hline
 Conv41 & Convolution & 3x3 / 1 & 13x13x128 \\ \hline
 Conv42 & Convolution & 3x3 / 1 & 13x13x256 \\ \hline
 Pool4 & MaxPooling & 2x2 / 2 & 7x7x256 \\ \hline
 Conv51 & Convolution & 3x3 / 1 & 7x7160 \\ \hline
 Conv52 & Convolution & 3x3 / 1 & 7x7x320 \\ \hline
 Pool5 & AvgPooling & 7x7 / 1 & 1x1x320 \\ \hline
 \end{tabular}
 \end{adjustbox}
 \vspace{1mm}
 \caption{Face feature extraction network structure.}
 \label{tab:casia_network}
 \end{minipage}
 \hfill
 \begin{minipage}{0.52\textwidth}
 \centering
 \begin{adjustbox}{max width=0.99\textwidth}
 \begin{tabular}{| l | l | c | c |}
 \hline
 Name & Type & Filter size / Stride & Output size \\
 \hline\hline
 Conv1 & Convolution & 5x5 / 2 & 28x28x32 \\
 \hline
 Pool1 & MaxPooling & 2x2 / 1 & 14x14x32 \\
 \hline
 Conv2 & Convolution & 5x5 / 1 & 14x14x32 \\
 \hline
 Pool2 & MaxPooling & 2x2 / 1 & 7x7x32 \\
 \hline
 Dropout3 & Dropout & & 7x7x32 \\
 \hline
 FC3 & FullyConnected & & 1x1x256 \\
 \hline
 Dropout4 & Dropout & & 1x1x256 \\
 \hline
 \end{tabular}
 \end{adjustbox}
 \vspace{1mm}
 \caption{MNIST feature extraction network structure.}
 \label{tab:mnist-net}	
 
 \begin{adjustbox}{max width=0.99\textwidth}
 \begin{tabular}{| l | l | c | c |}
 \hline
 Name & Type & Filter size / Stride & Output size \\
 \hline\hline
 Conv1 & Convolution & 5x5 / 1 & 32x32x32 \\
 \hline
 Pool1 & MaxPooling & 3x3 / 2 & 16x16x32 \\
 \hline
 Conv2 & Convolution & 5x5 / 1 & 16x16x32 \\
 \hline
 Pool2 & MaxPooling & 3x3 / 2 & 8x8x32 \\
 \hline
 Conv3 & Convolution & 5x5 / 1 & 8x8x64 \\
 \hline
 Pool3 & AvgPooling & 3x3 / 2 & 4x4x64 \\
 \hline
 Dropout3 & Dropout & & 4x4x64 \\
 \hline
 FC3 & FullyConnected & & 1x1x64 \\
 \hline
 \end{tabular}	 
 \end{adjustbox}
 \vspace{1mm}
 \caption{CIFAR10 feature extraction network structure.}
 \label{tab:cifar-net}
 \end{minipage}
\end{table}
 \vspace{-8mm}

%% file: 05_conclusions.tex
\section{Discussion}

In this work, we tackle the challenging problem of hashing image sets for scalable matching.
To do that, we introduce a set hashing deep network which takes an image set as input, and process it with image feature extraction, computing set feature and finally convert it to a single binary code. The network is learned from end-to-end where triplet loss is used to optimize the hashing output and feature learning. 
Extensive experiments are conducted showing our set based hashing is superior to single image based hashing which is also trained using deep models. Our method is also able to achieve similar retrieval performance with state-of-the-art set matching algorithms which does not have a compact representation as binary codes and use more complicated class-specific models. Further experiments show our approach is able to improve with larger set size and the set features we used are complementary and work better when used together. 

Set hashing is a valid solution to a real-world problem when dealing with large collection of images. Our future effort will be dedicated to the design of a network architecture which can potentially learn set features and domain specific set data generation.